\documentclass[letterpaper]{article} % DO NOT CHANGE THIS
\usepackage{aaai2027}  % DO NOT CHANGE THIS
\usepackage[hyphens]{url}  % DO NOT CHANGE THIS
\usepackage{graphicx} % DO NOT CHANGE THIS
\usepackage{natbib}  % DO NOT CHANGE THIS AND DO NOT ADD ANY OPTIONS TO IT
\usepackage{caption} % DO NOT CHANGE THIS AND DO NOT ADD ANY OPTIONS TO IT
\usepackage{algorithm}
\usepackage{algorithmic}
\usepackage{booktabs}
\usepackage{multirow}
\usepackage{amsmath}
\usepackage{amssymb}
\usepackage{newfloat}
\usepackage{listings}
\DeclareCaptionStyle{ruled}{labelfont=normalfont,labelsep=colon,strut=off} % DO NOT CHANGE THIS
\floatstyle{ruled}
\newfloat{listing}{tb}{lst}{}
\floatname{listing}{Listing}

\usepackage{booktabs}

\title{RRM: Experience-Driven Reflective Retrieval Memory for
Long-Horizon Multimodal Reasoning}

\author{
    Jingxiang Fan\textsuperscript{\rm 1},
    Junbao Zhuo\textsuperscript{\rm 1},
    Bochao Zou\textsuperscript{\rm 1}\corresponding
}

\affiliations{
    \textsuperscript{\rm 1}University of Science and Technology Beijing\\
    30 Xueyuan Road, Haidian District, Beijing 100083, China\\
    u202242070@xs.ustb.edu.cn,
    junbaozhuo@ustb.edu.cn,
    zoubochao@ustb.edu.cn
}

\begin{document}

\maketitle

\begin{abstract}
Existing multimodal long-term memory agents use external memory to overcome the limited context available for long videos. However, most methods emphasize what to store rather than how stored memory should be retrieved. When retrieval becomes inaccurate or repeatedly fails to obtain useful evidence, existing agents lack mechanisms to diagnose failures from previous task trajectories and adapt future search strategies.
We introduce \textbf{Reflective Retrieval Memory (RRM)}, a reflective memory framework for long-horizon multimodal reasoning. RRM augments an entity-centric multimodal memory graph with reflective experience memory, which distills transferable procedural retrieval knowledge from historical task trajectories. Unlike episodic and semantic memories that preserve factual evidence from the current video, reflective experience memory captures reusable search strategies across tasks. RRM converts retrieved experiences into query-level guidance, while answer generation remains conditioned only on factual evidence newly retrieved from the current video. A lifecycle management mechanism further regulates experience memory through usage frequency, reuse feedback, and temporal decay, thereby reducing redundancy and noise.
RRM consistently outperforms previous state-of-the-art approaches on M3-Bench-Robot, M3-Bench-Web, and Video-MME-Long, demonstrating the effectiveness of reflective retrieval memory for long-horizon multimodal reasoning. 
% RRM outperforms previous state-of-the-art approaches by \textbf{9.1\%, 5.8\%, and 7.4\%} on M3-Bench-Robot, M3-Bench-Web, and Video-MME-Long, respectively. 
\end{abstract}
\begin{figure*}[t]
    \centering
    \includegraphics[width=\textwidth]{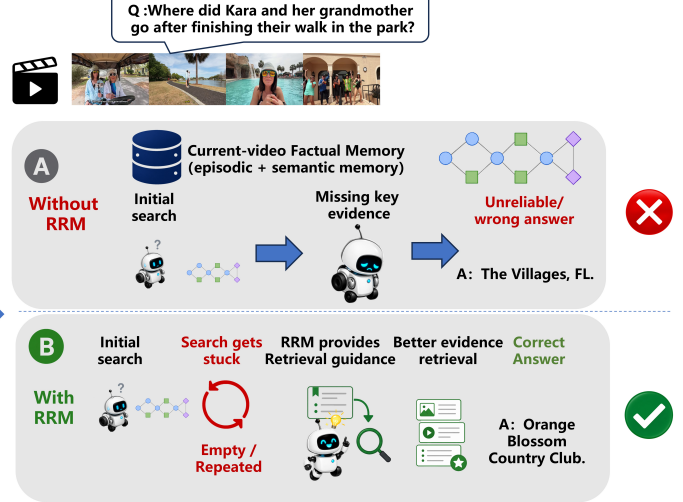}
    \caption{Motivation and retrieval workflow of RRM. Procedural
retrieval experience distilled from prior successful and failed tasks
provides query-level guidance to resolve retrieval anomalies and
acquire missing evidence from current-video memory.}
    \label{fig:overview}
\end{figure*}
\section{Introduction}

Multimodal large language models (MLLMs) have demonstrated
strong perception and reasoning capabilities in video understanding,
cross-modal question answering, and embodied reasoning~\citep{bai2025qwen25vl,xu2025qwen25omni,reid2024gemini15,
openai2024gpt4o}. However,
continuous long-horizon scenarios impose substantially greater
demands on these models. In applications such as household robotics
and persistent environmental perception, an agent must preserve
visual, semantic, and temporal information over continuous video
streams while accurately locating evidence relevant to the current
question across multiple tasks. Long videos, however, contain
substantial redundancy, with critical evidence sparsely distributed
across distant temporal segments~\citep{song2024moviechat,he2024malmm,fu2025videomme}. Meanwhile, despite recent extensions of nominal context windows,
existing models still exhibit limited effective context capacity and
struggle to reliably locate and use relevant evidence over long inputs,
particularly in long-form video understanding
\cite{liu-etal-2024-lost,hsieh2024ruler,wu2024longvideobench}. Consequently, the central bottleneck in long-horizon
multimodal reasoning is not merely how to retain more historical
information, but how to reliably retrieve supporting evidence from
large-scale memory and further improve future evidence acquisition
using experience from previous tasks.

Learning from historical task trajectories has become an important
approach to continual adaptation in language agents. Existing methods
generate verbal reflections from feedback, extract reusable insights
from successful and failed experiences, or induce reusable workflows
to guide subsequent reasoning and action generation
~\citep{shinn2023reflexion,zhao2024expel,wang2025awm,cao2026reme}.
However, a long-video reasoning trajectory contains two distinct kinds of information. One consists of task-specific facts, such as people, locations, and events; the other captures how relevant evidence was retrieved. Directly reusing the complete trajectory may
introduce irrelevant or conflicting historical context, potentially
directing the agent toward entities absent from the current video or
encouraging it to answer from previous-task information rather than
retrieve new evidence
~\citep{shi2023distracted,hong2024gullible}.
The transferable object should therefore be the retrieval procedure,
rather than the historical answer or its supporting facts.

As illustrated in Figure~\ref{fig:overview}, RRM extracts reusable
retrieval experience from historical task trajectories rather than
preserving task-specific answers. For example, in a previous task
asking where Blondie went after exploring the ancient city of
Dukezong, the agent successfully located the subsequent destination,
Pudacuo Park. The transferable information in this trajectory is not
the destination itself, but the underlying retrieval strategy:
localizing a known event and searching subsequent temporal segments
for evidence of an event transition. Similarly, failed trajectories
can reveal diagnostic information about ineffective queries, repeated
searches, or insufficient evidence coverage. By abstracting successful
and failed trajectories into retrieval strategies, RRM can redirect
evidence search in future tasks without introducing facts from
historical videos into current-video reasoning. RRM follows a predict-then-update protocol. For each task, the prediction is finalized and recorded before post-task feedback becomes available. The resulting experience can therefore affect only subsequent tasks, establishing temporal separation between prediction and experience update.

Based on the above observations, we propose \textbf{Reflective Retrieval
Memory (RRM)}, a framework for long-horizon multimodal reasoning.
Built upon the graph-structured long-term memory of
M3-Agent~\citep{long2026m3agent}, RRM augments the episodic and
semantic memories that preserve factual evidence from the current
video with a reflective experience memory continuously distilled from
historical task trajectories. Through reflection on successful and
failed retrieval trajectories, RRM abstracts transferable evidence
requirements, query patterns, failure causes, and correction strategies
rather than retaining historical answers or video-specific facts. To
prevent task-specific historical information from interfering with
current reasoning, RRM models experience as a query-level retrieval
control signal. Retrieved experiences are filtered by an experience
selector, transformed into neutral retrieval focuses, and used solely
to construct subsequent memory queries. They are never directly
inserted into the answer-generation context. In this way, the agent can
reuse procedural retrieval knowledge across tasks while ensuring that
its final prediction remains grounded in factual evidence retrieved
from the current video.

As tasks are continuously processed, the reflective experience memory
may accumulate redundant, outdated, or ineffective entries. We
therefore introduce an online lifecycle management mechanism that
dynamically estimates experience utility according to actual reuse,
task feedback, and temporal decay. Similar experiences are consolidated,
while ineffective entries are pruned under explicit capacity control,
allowing the experience memory to remain compact and reliable over
long task streams.

Our contributions are summarized as follows:
\begin{itemize}
    \item We propose RRM, a framework that learns transferable procedural retrieval knowledge from historical task trajectories to improve cross-task retrieval in long-horizon multimodal reasoning. 
    \item We introduce a query-level experience reuse mechanism that transforms historical experiences into retrieval control signals and uses them only for query construction, preventing task-specific historical information from directly affecting answer generation. 
    \item We conduct extensive experiments on M3-Bench-Robot, M3-Bench-Web, and Video-MME-Long, where RRM consistently outperforms strong baselines with improvements of 9.1\%, 5.8\%, and 7.4\%, respectively. 
\end{itemize}

\section{Related Work}
\begin{figure*}[t]
    \centering
    \includegraphics[width=\textwidth]{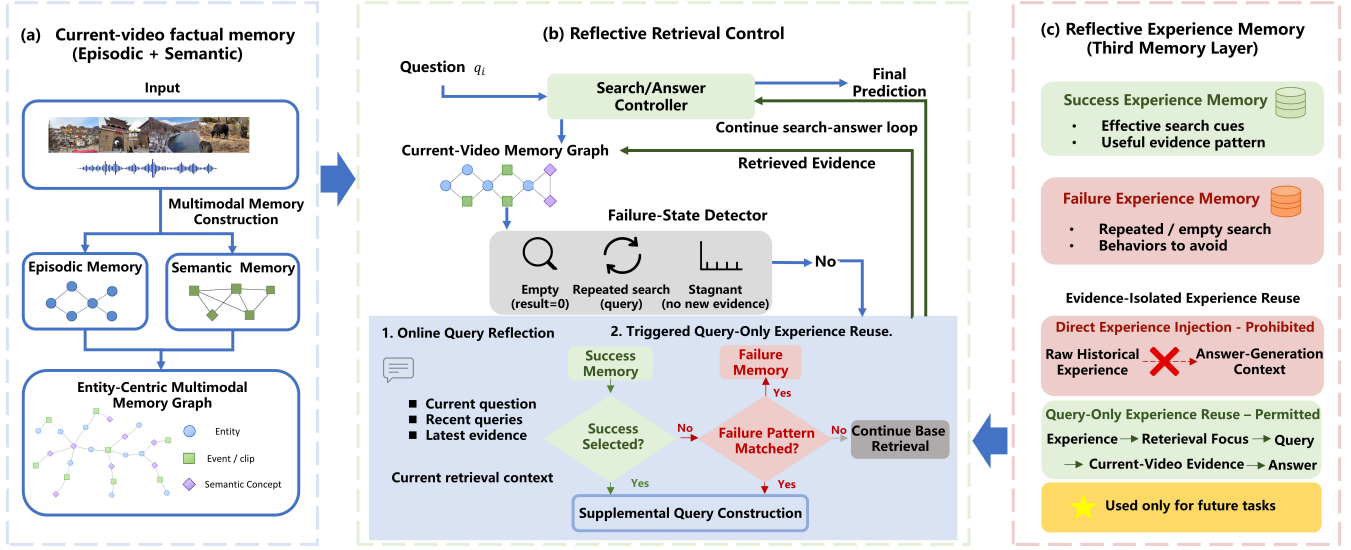}
    \caption{Overall architecture of RRM: (a) current-video factual
memory, (b) reflective retrieval control with Online Query Reflection
and triggered query-only experience reuse, and (c) reflective
experience memory with evidence-isolated reuse.}
    \label{fig:framework}
\end{figure*}
\textbf{Multimodal Long-Video Understanding and
Memory-Augmented Agents.}\ 
Critical evidence in long videos is often sparsely distributed across
different temporal segments. Existing studies primarily reduce the
cost of long-context processing through video representation
compression and active evidence retrieval. MovieChat
~\citep{song2024moviechat} compresses long-video representations into
sparse memory, while MA-LMM~\citep{he2024malmm} processes videos
online and maintains historical visual information in a memory bank.
VideoAgent~\citep{wang2024videoagent} formulates long-video question
answering as an iterative search process. M3-Agent
~\citep{long2026m3agent} further constructs entity-centric episodic and
semantic memories and retrieves task-relevant evidence through a
Search--Answer controller, whereas WorldMM~\citep{yeo2026worldmm}
explores dynamic access to memories across different modalities and
temporal scales. These methods substantially improve the storage and accessibility of
long-video information. However, their retrieval mechanisms are
primarily designed for evidence acquisition within the current video
or task, with limited use of historical task trajectories to learn
retrieval strategies that can be reused across tasks. In contrast, RRM
preserves the factual memory of the current video while introducing an
independent reflective experience memory that continually accumulates
and reuses procedural retrieval knowledge.

\textbf{Experience-Based and Self-Evolving Agent Memory.}
Experience-driven agents transform historical trajectories and
environmental feedback into reusable reflections, experiences, or
workflows. Reflexion~\citep{shinn2023reflexion} generates
natural-language reflections from task feedback,
ExpeL~\citep{zhao2024expel} extracts reusable experience from
successful and failed trajectories, and Agent Workflow Memory
~\citep{wang2025awm} induces reusable workflows from historical
execution trajectories. ReMe~\citep{cao2026reme} further studies the
reuse, updating, and continual evolution of structured experience.
RRM differs from these approaches primarily in where experience is
introduced into the reasoning process. Rather than directly exposing
historical experience to the reasoning or answer-generation context,
RRM converts it into query-level retrieval guidance. The final answer
is then generated exclusively from factual evidence retrieved from the
current-video memory. This design reduces the influence of irrelevant
or conflicting historical context, and is empirically examined
in Figure~\ref{fig:reuse_efficiency}(a).

\section{Method}

The core idea of RRM is to distill transferable procedural retrieval
experience from historical task trajectories to improve evidence
acquisition for long-horizon multimodal reasoning. RRM is reflective because it examines how retrieval trajectories succeed or fail and converts the resulting diagnoses into reusable guidance for subsequent evidence acquisition.  RRM retains the current-video factual memory
(Figure~2(a)) and extends its Search--Answer process with
Online Query Reflection and anomaly-triggered experience reuse
(Figure~2(b) and (c)). Retrieved experiences are
converted into query-level guidance, while final predictions remain
grounded only in factual evidence retrieved from the current video.
After task completion, new experiences are extracted from the retrieval
trajectory and maintained through an online lifecycle mechanism,
forming a cross-task retrieval optimization loop.

\subsection{Preliminary}

RRM builds on the memory construction and control reasoning pipeline of M3-Agent (Long et al. 2026). Continuous audiovisual streams are organized into an entity-centric multimodal graph \(G_i\), whose episodic memory preserves temporally grounded event observations and whose semantic memory abstracts entity attributes and relations. For task \(i\), the controller issues query \(x_{i,t}\) at retrieval step \(t\) and obtains

\begin{equation}
E_{i,t}=\mathrm{Search}(\mathcal{G}_{i},x_{i,t}).
\end{equation}

where \(E_{i,t}\) denotes the retrieved evidence. The Search--Answer process iterates under the controller policy until an answer is produced or the retrieval budget is exhausted. RRM retains this factual retrieval backbone and augments query construction only when retrieval failures are detected, leveraging reflective experiences from previous task trajectories to refine future evidence acquisition.

\subsection{Three-Layer Memory Architecture}

% RRM organizes long-term memory into three layers: episodic memory,
%semantic memory, and reflective experience memory. This section
%describes the information sources, represented content, and functional
%roles of each memory type in detail.
As shown in Figure~\ref{fig:framework}(a) and (c), RRM organizes
long-term memory into three layers: episodic memory, semantic memory,
and reflective experience memory.

\paragraph{Current-Video Factual Memory.}
As illustrated in Figure~\ref{fig:framework}(a), episodic and
semantic memories are constructed exclusively from multimodal
observations of the current video and jointly form the factual memory
graph $\mathcal{G}_i$. Episodic memory uses temporally grounded video
segments and events as its basic units, whereas semantic memory
abstracts higher-level entity attributes and relations. Together, they
organize the entities, events, and relations of the current video into
an entity-centric multimodal memory graph for evidence retrieval.

\paragraph{Reflective Experience Memory.}
As shown in Figure~\ref{fig:framework}(c), reflective experience
memory forms an independent third layer that compresses completed
retrieval trajectories into procedural knowledge for future tasks.
Unlike episodic and semantic memories constructed from the current
video, it is distilled from historical retrieval trajectories and task
feedback. It stores transferable evidence requirements, query patterns,
retrieval failure modes, and query-adjustment strategies rather than
specific people, events, answers, or conclusions from historical
videos. It therefore describes how relevant evidence should be
acquired, rather than what occurred in previous videos.

To maintain a clear boundary between factual information and
procedural experience, reflective experience memory is kept independent
of the current-video factual graph. Historical experiences are never
written into episodic or semantic memory and are used only as
query-level retrieval control signals. This separation enables
cross-task transfer of retrieval strategies while preventing historical
entities, answers, or erroneous conclusions from becoming factual
input to current-video reasoning. For task $i$, the pre-feedback
trajectory is
\begin{equation}
\tau_i^{\mathrm{pre}}
=
\left(
q_i,
\left[(x_{i,t}, E_{i,t})\right]_{t=1}^{T_i},
\hat{y}_i
\right),
\end{equation}
where $\tau_i^{\mathrm{pre}}$ denotes the trajectory recorded before
post-task feedback, $q_i$ is the question for task $i$, and $T_i$ is
the total number of retrieval steps. Following the notation introduced
above, $\left[(x_{i,t},E_{i,t})\right]_{t=1}^{T_i}$ is the ordered
sequence of retrieval queries and corresponding evidence, and
$\hat{y}_i$ is the final prediction fixed before any post-task
feedback becomes available.

At the beginning of each independent evaluation run, the reflective
experience memory is initialized as empty. We process questions in
mini-batches of 64 and assign all questions associated with the same
video to the same mini-batch. All predictions within a mini-batch are
finalized before any ground-truth feedback is revealed. An LLM-based
evaluator then compares each prediction $\hat{y}_i$ with the
ground-truth answer $y_i^{\star}$ and produces a binary success signal
$s_i$.

For successful tasks, the experience extractor summarizes the
retrieval trajectory without access to the ground-truth answer. For
failed tasks, it additionally receives $y_i^{\star}$ and $\hat{y}_i$
to generate a corrective procedural retrieval strategy. Extracted
experiences are written to memory only after the entire mini-batch has
been scored. They can therefore affect only subsequent mini-batches.
Because questions from the same video are never split across
mini-batches, an experience cannot influence another question from its
source video. This procedure defines a mini-batch online adaptation
protocol with delayed ground-truth feedback.

Corresponding to the two experience banks in
Figure~\ref{fig:framework}(c), RRM maintains separate memories for
successful and failed tasks:
\begin{equation}
\mathcal{M}^{\mathrm{ref}}
=
\left\{
\mathcal{M}^{+},
\mathcal{M}^{-}
\right\},
\end{equation}
where $\mathcal{M}^{\mathrm{ref}}$ denotes the overall reflective
experience memory, consisting of a successful experience memory
$\mathcal{M}^{+}$ and a failure experience memory $\mathcal{M}^{-}$.
The superscripts $+$ and $-$ indicate that the stored experiences are
extracted from successful and failed task trajectories, respectively.
$\mathcal{M}^{+}$ stores effective evidence requirements and
query-organization patterns distilled from successful trajectories.
In contrast, $\mathcal{M}^{-}$ stores retrieval failure patterns and
the corresponding query-adjustment strategies derived from failed
trajectories. The two memories are maintained independently because
their contents have different levels of reliability.

Each experience is represented as a structured procedural record that
specifies the task type, applicability conditions, required evidence,
retrieval strategy, undesirable retrieval behaviors, and
query-adjustment patterns. Rather than preserving complete historical
trajectories, the extraction process removes historical answers,
option labels, and factual conclusions specific to the source task.
It then abstracts the remaining information into retrieval patterns
that do not depend on entities from the source video.

Only records satisfying the predefined structural and content
constraints are retained. The complete extraction prompts, record
schemas, filtering constraints, model configurations, and relevant
hyperparameters are provided in
Appendix.

Each procedural record is stored together with its semantic
representation to support subsequent experience retrieval. At
retrieval time, RRM excludes all experiences derived from the current
video. A task can therefore retrieve only procedural experiences
obtained from other videos. Together with the removal of historical
answers and entities, this cross-video isolation reduces the risk of
introducing task-specific historical information into current-video
reasoning. The next subsection describes the stricter extraction and
reuse constraints applied to experiences derived from failed tasks.

\subsubsection{Failure-Aware Experience Modeling}

Compared with successful trajectories, failed trajectories contain
more complex and less reliable information. Retrieval failures may
arise from overly broad queries, incorrect entity localization,
missing critical evidence, or misaligned cross-modal information.
Therefore, directly storing and reusing complete failed trajectories
is unreliable, as unfiltered failure information may cause the agent
to repeat previous errors in subsequent tasks.

RRM does not treat failed trajectories themselves as reusable
experiences. Instead, it extracts corrective procedural knowledge
related to the retrieval process. Failure experience extraction focuses
on three aspects: retrieval anomalies observed during the search,
evidence types missing for the current question, and directions for
adjusting future queries. For example, when multiple retrieval rounds
repeatedly return the same video segments, the extracted experience
should recommend changing the retrieval perspective or introducing
additional constraints, rather than preserving the original queries
and erroneous conclusions.

Corrective feedback obtained after task completion is used only as a
diagnostic signal to identify failure causes and evidence not covered
by the retrieval trajectory. It is not stored as factual content in
reflective experience memory.

Although experiences extracted from successful and failed tasks share the same storage structure, they differ in reliability.  A successful trajectory
validates its retrieval strategy through a correct task outcome,
whereas a failed trajectory provides only a diagnostic hypothesis
that may not generalize. RRM therefore maintains the two types of
experience separately and adopts an asymmetric selection and reuse
mechanism during inference, imposing stricter applicability constraints
on failure experiences.

\subsection{Reflective Retrieval Control}

\paragraph{Online Query Reflection.}

Before consulting cross-task experience, RRM performs Online Query
Reflection (OQR) using only the current task state. OQR addresses
within-task retrieval failures that the original Search--Answer
controller does not explicitly diagnose. As illustrated in
Figure~\ref{fig:framework}(b), a rule-based Failure-State Detector
monitors the ongoing Search--Answer process and determines whether
the anomaly trigger should be activated. The anomaly trigger is
activated when the latest query returns no valid evidence, recent
queries are identical or highly similar, or consecutive retrieval
rounds obtain no new evidence.

At retrieval step $t$, OQR examines the current question, previously
executed queries, and returned evidence. It then generates a neutral
local repair query $\widetilde{x}_{i,t}$ for subsequent retrieval
without using historical experience. OQR is therefore distinct from
cross-task experience reuse: it corrects the current retrieval
trajectory using only current-task information and corresponds to
component A in Table~2.

\paragraph{Triggered Query-Only Experience Reuse.}

Under the same anomaly trigger, RRM may additionally consult reflective
experience memory for cross-task guidance. It constructs the current
retrieval state from the question, anomaly type, and search history,
and retrieves candidates separately from $\mathcal{M}^{+}$ and
$\mathcal{M}^{-}$. An experience selector evaluates each candidate
according to its stored applicability conditions, evidence requirements,
and failure modes. Applicable successful experiences are prioritized.
A failure experience is considered only when no applicable successful
experience is available and its diagnostic conditions match the current
retrieval anomaly.

The selected experience is converted into the search direction and
evidence constraints needed for the current task. Historical answers,
source-task entities, and factual conclusions are excluded, producing
an experience-derived retrieval focus $f_{i,t}^{\mathrm{ref}}$. OQR
and reflective experience reuse therefore provide complementary
signals: $\widetilde{x}_{i,t}$ repairs the current search using
within-task context, whereas $f_{i,t}^{\mathrm{ref}}$ transfers a
retrieval procedure learned from previous tasks.

The query-construction stage combines the local repair query, the
experience-derived retrieval focus, and the auxiliary retrieval keys
retained from the baseline. Let $\mathcal{Q}_{i,\leq t}$ denote the
set of queries already executed for the current task, and let
$\mathcal{K}_{i,t}^{\mathrm{base}}$ denote the entity- or
option-related auxiliary retrieval keys retained from the baseline.
The supplemental query set is defined as

\begin{equation}
\mathcal{C}_{i,t}^{\mathrm{ref}}
=
\operatorname{Dedup}_{L}
\left(
\left(
\left\{
\widetilde{x}_{i,t},
f_{i,t}^{\mathrm{ref}}
\right\}
\cup
\mathcal{K}_{i,t}^{\mathrm{base}}
\right)
\setminus
\mathcal{Q}_{i,\leq t}
\right),
\label{eq:supplemental_queries}
\end{equation}

where $\operatorname{Dedup}_{L}$ removes empty and duplicate queries
while retaining at most $L$ candidates. The set difference excludes
queries already executed for the current task, preventing redundant
repetition of previously explored retrieval directions.

Neither the original experience record nor its historical task
information is included in the answer-generation context. Historical
experience can affect the final prediction only by changing which
evidence is newly retrieved from the current-video factual memory.
The answer generator therefore receives current-video evidence rather
than historical factual content.
\subsection{Online Memory Lifecycle Management}

RRM maintains experience utility using actual reuse, task feedback, and temporal decay. Usage and reuse feedback are updated only when an experience-guided query is executed and successfully retrieves new evidence from the current video. Unused experiences gradually decrease in priority through temporal decay. To limit growth, RRM merges highly similar successful experiences and aggregates their feedback, but keeps failure experiences separate because they may encode different diagnostic conditions. It periodically prunes unused or low-utility experiences while protecting newly created entries during a cold-start period. These operations control redundancy without changing the separation between procedural experience and factual memory.

\section{Experiments}

\begin{table*}[t]
\centering
\caption{
Comparison with state-of-the-art methods on long-video multimodal
reasoning benchmarks. ME, MH, CM, PU, and GK denote Multi-Event,
Multi-Hop, Cross-Modal, Person Understanding, and General Knowledge,
respectively. The best and second-best results are highlighted in
\textbf{bold} and \underline{underlined}, respectively.
}
\label{tab:main_results}

\resizebox{\textwidth}{!}{
\begin{tabular}{l|cccccc|cccccc|c}
\toprule
\multirow{2}{*}{Method}
& \multicolumn{6}{c|}{M3-Bench-Robot}
& \multicolumn{6}{c|}{M3-Bench-Web}
& \multirow{2}{*}{Video-MME-Long} \\
\cmidrule(lr){2-7}
\cmidrule(lr){8-13}

& ME & MH & CM & PU & GK & ALL
& ME & MH & CM & PU & GK & ALL
& \\
\midrule

\multicolumn{14}{c}{
\textbf{General Multimodal Large Language Models}
} \\
\midrule

Qwen2.5-Omni-7B~\citep{xu2025qwen25omni}
& 2.1 & 1.4 & 1.5 & 1.5 & 2.1 & 2.0
& 8.9 & 8.8 & 13.7 & 10.8 & 14.1 & 11.3
& 42.2 \\

Qwen2.5-VL-7B~\citep{bai2025qwen25vl}
& 2.9 & 3.8 & 3.6 & 4.6 & 3.4 & 3.4
& 11.9 & 10.5 & 13.4 & 14.0 & 20.9 & 14.9
& 46.9 \\

Gemini-1.5-Pro~\citep{reid2024gemini15}
& 6.5 & 7.5 & 8.0 & 9.7 & 7.6 & 8.0
& 18.0 & 17.9 & 23.8 & 23.1 & 28.7 & 23.2
& 38.0 \\

GPT-4o~\citep{openai2024gpt4o}
& 9.3 & 9.0 & 8.4 & 10.2 & 7.3 & 8.5
& 21.3 & 21.9 & 30.9 & 27.1 & 39.6 & 28.7
& 38.8 \\

\midrule
\multicolumn{14}{c}{
\textbf{Online Long-Video Understanding Methods}
} \\
\midrule

MovieChat~\citep{song2024moviechat}
& 13.3 & 9.8 & 12.2 & 15.7 & 7.0 & 11.2
& 12.2 & 6.6 & 12.5 & 17.4 & 11.1 & 12.6
& 19.4 \\

MA-LMM~\citep{he2024malmm}
& 25.6 & 23.4 & 22.7 & 39.1 & 14.4 & 24.4
& 26.8 & 10.5 & 22.4 & 39.3 & 15.8 & 24.3
& 17.3 \\

Flash-VStream~\citep{zhang2025flashvstream}
& 21.6 & 19.4 & 19.3 & 24.3 & 14.1 & 19.4
& 24.5 & 10.3 & 24.6 & 32.5 & 20.2 & 23.6
& 25.0 \\

VideoChat-Online~\citep{huang2025videochatonline}
& 31.7 & 24.7 & 30.5 & 43.1 & 17.1 & 29.9
& 34.3 & 18.9 & 34.1 & 48.3 & 23.1 & 32.7
& 45.9 \\

TimeChat-Online~\citep{yao2025timechatonline}
& 35.6 & 29.4 & 31.5 & 44.3 & 22.3 & 33.6
& 38.8 & 22.8 & 38.4 & 51.6 & 28.0 & 36.6
& 49.2 \\

StreamForest~\citep{zeng2025streamforest}
& 33.4 & 29.4 & 31.3 & 44.3 & 19.0 & 32.1
& 37.9 & 22.3 & 39.9 & 51.7 & 27.6 & 36.5
& 51.9 \\

\midrule
\multicolumn{14}{c}{
\textbf{Agent-Based Long-Term Memory Methods}
} \\
\midrule

Gemini-Agent~\citep{long2026m3agent}
& 15.8 & 17.1 & 15.3 & 20.0 & 15.5 & 16.9
& 29.3 & 20.9 & 33.8 & 34.6 & 45.0 & 34.1
& 55.1 \\

Gemini-GPT4o-Hybrid~\citep{long2026m3agent}
& 21.3 & 25.5 & 22.7 & 28.8 & 23.1 & 24.0
& 35.9 & 26.2 & 37.6 & 43.8 & 52.2 & 41.2
& 56.5 \\

M3-Agent~\citep{long2026m3agent}
& 32.0 & 31.8 & 31.1 & 42.2 & 19.9 & 30.5
& 43.4 & 28.3 & 45.0 & 59.0 & 54.1 & 48.6
& 56.4 \\

NS-Mem~\citep{jiang2026nsmem}
& \underline{36.2}
& 31.5
& 33.8
& \underline{45.7}
& 26.4
& \underline{34.7}
& \textbf{54.2}
& \underline{34.6}
& \underline{47.4}
& 60.1
& \underline{59.7}
& \underline{53.6}
& -- \\

StreamMeCo~\citep{wang2026streammeco}
& 35.7
& \underline{32.9}
& \underline{35.5}
& 44.0
& \underline{27.5}
& 34.6
& 46.4
& 32.2
& 46.5
& \underline{61.8}
& 55.8
& 50.7
& \underline{56.6} \\

\midrule

\textbf{RRM}
& \textbf{40.1}
& \textbf{37.7}
& \textbf{38.9}
& \textbf{52.0}
& \textbf{29.4}
& \textbf{39.6}
& \underline{50.0}
& \textbf{37.0}
& \textbf{50.6}
& \textbf{62.0}
& \textbf{61.0}
& \textbf{54.4}
& \textbf{63.8} \\

\bottomrule
\end{tabular}
}
\end{table*}

\paragraph{Evaluation Protocol.}
We evaluate RRM using an online mini-batch adaptation protocol with
delayed ground-truth feedback, following the streaming test-time
learning setting of Evo-Memory~\cite{wei2025evomemory}. Questions associated with the same source video are kept within a single mini-batch. We finalize all predictions in each mini-batch before revealing ground-truth feedback or updating reflective experience memory. Experiences extracted from one mini-batch can therefore affect only subsequent mini-batches, never another question from the same source video. This protocol enables forward-only online adaptation while preventing within-video feedback leakage.

\subsection{Experimental Setup}

\paragraph{Datasets.}
We evaluate RRM on three long-video multimodal reasoning benchmarks:
M3-Bench-Robot, M3-Bench-Web~\citep{long2026m3agent}, and Video-MME-Long~\citep{fu2025videomme}.
M3-Bench-Robot contains 100 real-world videos captured from a
first-person robotic perspective, whereas M3-Bench-Web comprises
920 diverse open-domain videos collected from the Web. Both benchmarks
cover five reasoning categories: Multi-Event (ME), Multi-Hop (MH),
Cross-Modal (CM), Person Understanding (PU), and General Knowledge
(GK). We further evaluate RRM on Video-MME-Long to examine its
evidence retrieval and reasoning capabilities in extremely long-video
scenarios.

\paragraph{Baselines.}
We compare RRM against three groups of methods. The first group includes general-purpose multimodal large language
models: Qwen2.5-Omni-7B~\citep{xu2025qwen25omni},
Qwen2.5-VL-7B~\citep{bai2025qwen25vl},
Gemini-1.5-Pro~\citep{reid2024gemini15}, and
GPT-4o~\citep{openai2024gpt4o}.
The second group consists of online long-video understanding methods,
including MovieChat~\citep{song2024moviechat},
MA-LMM~\citep{he2024malmm},
Flash-VStream~\citep{zhang2025flashvstream},
VideoChat-Online~\citep{huang2025videochatonline},
TimeChat-Online~\citep{yao2025timechatonline}, and
StreamForest~\citep{zeng2025streamforest}.
The third group comprises agent-based long-term memory methods,
including Gemini-Agent and Gemini-GPT4o-Hybrid
~\citep{long2026m3agent}, M3-Agent~\citep{long2026m3agent},
NS-Mem~\citep{jiang2026nsmem}, and
StreamMeCo~\citep{wang2026streammeco}.
We use M3-Agent as the primary controlled baseline because RRM retains
its multimodal factual-memory construction pipeline and entity-centric
memory graph, while introducing reflective experience memory only
into the retrieval control process.

\paragraph{Implementation Details.}
To ensure a fair comparison, all RRM experiments employ the same
answer-generation model as the baseline. Query texts are encoded using
Qwen text-embedding-v3, and relevant video segments are retrieved from
the current-video memory graph for each query. The
controller performs at most five Search--Answer rounds before producing
the final prediction. Reflective experience memory is initialized as
empty and updated online throughout the task stream. At each retrieval
step, the system retrieves at most one successful experience and one
failure experience. After a task is completed and feedback becomes
available, the system extracts a successful or failure experience from
the corresponding retrieval trajectory according to the final
prediction outcome and stores it in the respective memory bank.
During experience retrieval, experiences derived from the current
video are excluded to enforce cross-video isolation. All experiments
are repeated three times, and the mean results are reported.
Complete implementation details, including the prompts, LLM configurations, structured experience schemas, applicability criteria, and lifecycle hyperparameters are provided in Appendix.

\subsection{Main Results}

As shown in Table~\ref{tab:main_results}, RRM achieves the best overall
performance across all three benchmarks. Compared with M3-Agent, which
shares the same factual-memory architecture and retrieval
infrastructure, RRM improves performance by 9.1, 5.8, and 7.4
percentage points on M3-Bench-Robot, M3-Bench-Web, and
Video-MME-Long, respectively. These gains demonstrate the effectiveness
of augmenting factual memory with cross-task procedural retrieval
experience.

At the category level, RRM achieves the best results across all five
reasoning categories on M3-Bench-Robot and ranks first on the MH, CM,
PU, and GK categories of M3-Bench-Web. Its slightly lower performance
than NS-Mem on the Web ME category suggests that aggregating evidence
across distributed events remains challenging. On Video-MME-Long,
RRM outperforms M3-Agent and the strongest competing method by 7.4
and 7.2 percentage points, respectively, showing that its benefits
extend to an extremely long-video benchmark with different video
content and evaluation settings. Overall, these results support the
effectiveness of extending long-term memory from factual storage to
retrieval-strategy learning.
% Required packages:
% \usepackage{booktabs}
% \usepackage{graphicx}
% \usepackage{amssymb}

\subsection{Ablation and Design Analysis}

%To systematically examine the performance gains and practical effectiveness of RRM, we conduct three complementary sets of ablation studies.

\subsubsection{Component Contributions}

To examine the incremental contribution of each component in RRM, we
conduct a progressive ablation study, with the results reported in
Table~\ref{tab:component_ablation}. The baseline retains the factual
memory and Search--Answer reasoning pipeline of M3-Agent, after which
Online Query Reflection, Reflective Experience Memory, and Lifecycle
Management are introduced sequentially.

Adding Online Query Reflection improves performance
by 4.4, 1.6, and 2.2 percentage points on M3-Bench-Robot,
M3-Bench-Web, and Video-MME-Long, respectively. OQR uses only the
current task state without historical experience. It diagnoses
retrieval anomalies from the current query, retrieved evidence, and
search trajectory, and accordingly revises subsequent queries.
Compared with the original Search--Answer controller, OQR explicitly
models the retrieval state and enables within-task query correction.
These gains indicate that adapting retrieval based solely on the
current search trajectory can already improve evidence acquisition
without cross-task experience transfer.

Building on this module, Reflective Experience Memory yields additional
gains of 3.1, 2.8, and 3.9 percentage points, respectively, indicating
that procedural knowledge distilled from historical successful and
failed trajectories provides useful cross-task retrieval guidance.
Lifecycle Management further improves performance by 1.6, 1.4, and
1.3 percentage points, suggesting that maintaining the experience
memory according to reuse feedback and temporal information helps
reduce interference from redundant or ineffective experiences. The
consistent incremental gains support the complementary roles of the
three components in within-task retrieval correction, cross-task
experience reuse, and long-term experience-quality maintenance.

\begin{table}[tb]
\centering
\caption{Ablation study of different components in RRM.}
\label{tab:component_ablation}

\setlength{\tabcolsep}{4pt}
\renewcommand{\arraystretch}{1.08}

\resizebox{\columnwidth}{!}{
\begin{tabular}{ccc|ccc}
\toprule
\multicolumn{3}{c|}{\textbf{Components}}
& \textbf{M3-Bench-Robot}
& \textbf{M3-Bench-Web}
& \textbf{Video-MME-Long} \\
\cmidrule(lr){1-3}
\cmidrule(lr){4-4}
\cmidrule(lr){5-5}
\cmidrule(lr){6-6}

\textbf{A}
& \textbf{B}
& \textbf{C}
& \textbf{Accuracy}
& \textbf{Accuracy}
& \textbf{Accuracy} \\
\midrule

--
& --
& --
& 30.5
& 48.6
& 56.4 \\

\checkmark
& --
& --
& 34.9
& 50.2
& 58.6 \\

\checkmark
& \checkmark
& --
& 38.0
& 53.0
& 62.5 \\

\checkmark
& \checkmark
& \checkmark
& \textbf{39.6}
& \textbf{54.4}
& \textbf{63.8} \\

\bottomrule
\end{tabular}
}

\vspace{1mm}
\begin{minipage}{\columnwidth}
\footnotesize
\textbf{A}: Online Query Reflection;
\textbf{B}: Reflective Experience Memory;
\textbf{C}: Lifecycle Management.
\end{minipage}

\end{table}

\subsubsection{Prompt-Level Injection vs. Query-Only Reuse}
\label{sec:reuse_strategy}

To examine where historical experience should be integrated, we compared prompt-level injection with query-only reuse under otherwise identical settings. Prompt-level injection appended the retrieved experience directly to the answer-generation context. In contrast, query-only reuse converted it into neutral guidance for subsequent query construction, while the answer generator received only factual evidence newly retrieved from the current-video memory graph.

As shown in Figure~\ref{fig:reuse_efficiency}(a), prompt-level injection improved over M3-Agent by 4.4, 4.7, and 2.6 percentage points on M3-Bench-Robot, M3-Bench-Web, and Video-MME-Long, respectively. Query-only reuse yielded additional gains of 4.7, 1.1, and 4.8 percentage points over prompt-level injection, with larger margins on M3-Bench-Robot and Video-MME-Long. This pattern supports using historical experience as a retrieval-control signal rather than exposing it directly to the answer generator. However, this comparison measures downstream performance and does not directly quantify how often prompt-level injection introduces task-specific historical content.
\subsubsection{Search Efficiency}

Beyond final task performance, we further evaluate the retrieval
efficiency of RRM. A retrieval round comprises the complete process in
which the controller selects a \textit{search} action, performs memory
retrieval, and receives the returned evidence; the final
\textit{answer} action is not counted as a retrieval round.

\begin{figure}[t]
    \centering
    \includegraphics[width=\columnwidth]{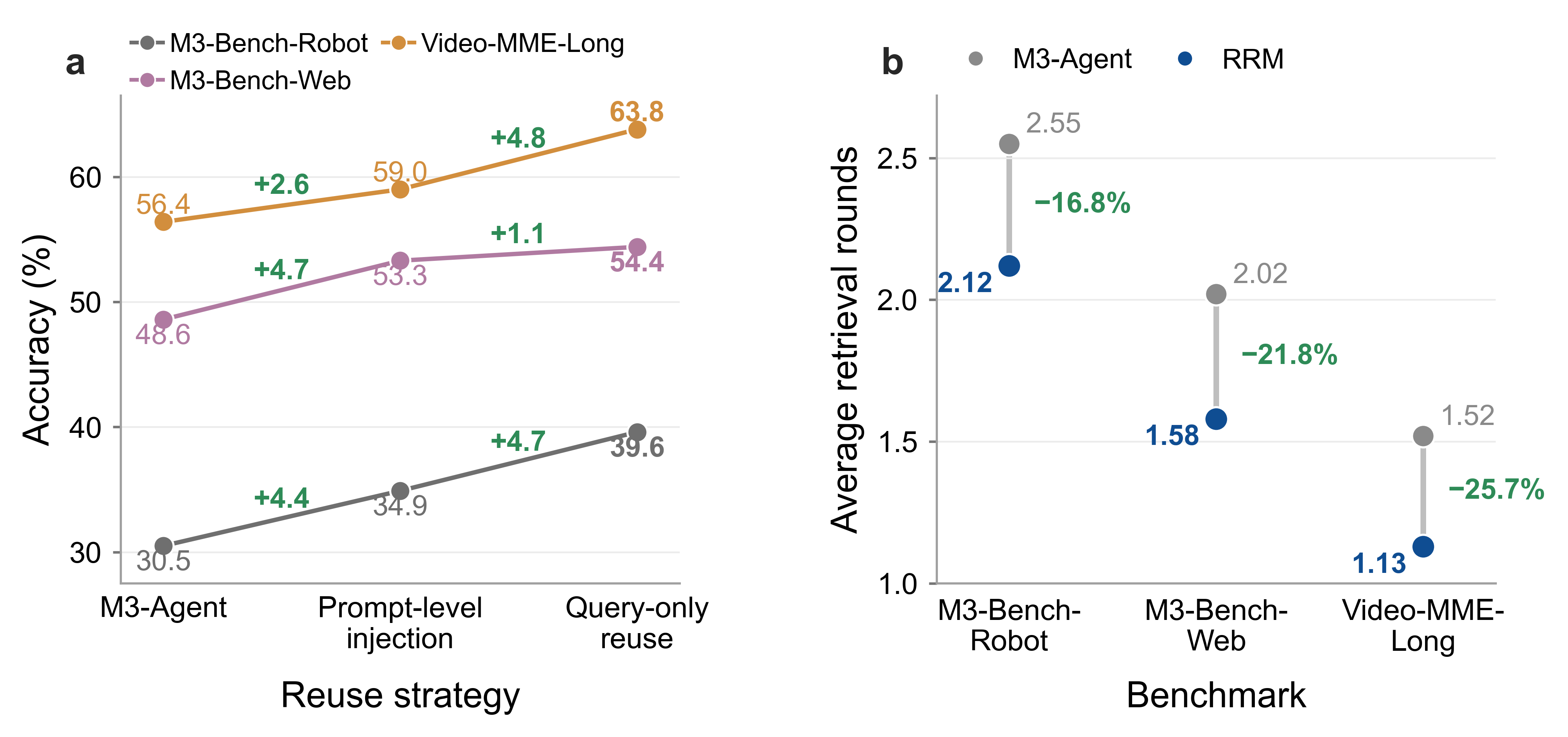}
    \caption{
    Experience reuse effectiveness and retrieval efficiency.
    (a) Accuracy comparison among M3-Agent, prompt-level injection,
    and query-only reuse.
    (b) Average retrieval rounds of M3-Agent and RRM.
    }
    \label{fig:reuse_efficiency}
\end{figure}

As shown in Figure~\ref{fig:reuse_efficiency}(b), RRM reduces the
average number of retrieval rounds from 2.55, 2.02, and 1.52 to 2.12,
1.58, and 1.13 on M3-Bench-Robot, M3-Bench-Web, and
Video-MME-Long, respectively, corresponding to reductions of
16.8\%, 21.8\%, and 25.7\%. Together with the main results in
Table~\ref{tab:main_results}, these findings show that RRM achieves
higher task performance with fewer retrieval iterations. Its gains
therefore do not arise from a larger retrieval budget, but from more
effective within-task query correction and cross-task experience reuse.
% \begin{table}[tb]
% \centering
% \caption{
% Experience reuse strategies and retrieval efficiency on three
% long-video benchmarks.
% }
% \label{tab:reuse_efficiency}
% \setlength{\tabcolsep}{5.5pt}
% \renewcommand{\arraystretch}{1.05}

% \begin{tabular}{lccc}
% \toprule
% Method / Strategy
% & Robot
% & Web
% & Video-MME-Long \\
% \midrule

% \multicolumn{4}{l}{
% \textit{(a) Accuracy (\%) of experience reuse strategies} $\uparrow$
% } \\
% Baseline
% & 30.5
% & 48.6
% & 56.4 \\
% Prompt-Level Injection
% & 34.9
% & 53.3
% & 59.0 \\
% Query-Only Reuse
% & \textbf{39.6}
% & \textbf{54.4}
% & \textbf{63.8} \\

% \midrule

% \multicolumn{4}{l}{
% \textit{(b) Average number of retrieval rounds} $\downarrow$
% } \\
% M3-Agent
% & 2.55
% & 2.02
% & 1.52 \\
% RRM
% & \textbf{2.12}
% & \textbf{1.58}
% & \textbf{1.13} \\
% Reduction
% & 16.8\%
% & 21.8\%
% & 25.7\% \\
% \bottomrule
% \end{tabular}
% \end{table}
\FloatBarrier

\section{Conclusion} This work addresses the difficulty of effectively retrieving critical evidence in long-video multimodal reasoning and proposes Reflective Retrieval Memory (RRM). RRM distills procedural retrieval experience from historical task trajectories and restricts its use to query-level control, thereby dynamically improving evidence acquisition from the current video without introducing historical factual information into answer generation. Experiments on three public benchmarks show that RRM outperforms the baseline by 9.1, 5.8, and 7.4 percentage points, respectively, while substantially reducing the average number of retrieval rounds. These results demonstrate that learning \emph{how to retrieve} can improve both the accuracy and efficiency of long-video multimodal reasoning.

% References and End of Paper
\bibliography{rrm_references}
\end{document}